\documentclass{article}

\usepackage{microtype}
\usepackage{graphicx}
\usepackage{subfigure}
\usepackage{wrapfig}

\usepackage{booktabs} 

\usepackage{hyperref}

\usepackage[accepted]{icml2020}
\usepackage{multirow}
\usepackage{url} 
\usepackage{xcolor}
\usepackage{bm}
\usepackage{amsmath}
\usepackage{amsfonts}
\usepackage{enumitem}

\icmltitlerunning{A Federated Learning Framework for Healthcare IoT devices}

\begin{document}

\twocolumn[
\icmltitle{A Federated Learning Framework for Healthcare IoT devices}



\icmlsetsymbol{equal}{*}

\begin{icmlauthorlist}
\icmlauthor{Binhang Yuan}{rice}
\icmlauthor{Song Ge}{uh}
\icmlauthor{Wenhui Xing}{prudence}
\end{icmlauthorlist}

\icmlaffiliation{rice}{Rice University, Houston, Texas, USA}
\icmlaffiliation{uh}{University of Houston-Downtown, Texas, USA}
\icmlaffiliation{prudence}{Prudence Medical Technologies Ltd., Shanghai, China}

\icmlcorrespondingauthor{Binhang Yuan}{by8@rice.edu}

\vskip 0.3in
]



\printAffiliationsAndNotice{}  

\begin{abstract}

The Internet of Things (IoT) revolution has shown potential to give rise to many medical applications with access to large volumes of healthcare data collected by IoT devices. 
However, the increasing demand for healthcare data privacy and security makes each IoT device an isolated island of data. 
Further, the limited computation and communication capacity of wearable healthcare devices restrict the application of vanilla federated learning. 
To this end, we propose an advanced federated learning framework to train deep neural networks, where the network is partitioned and allocated to IoT devices and a centralized server. 
Then most of the training computation is handled by the powerful server. 
The sparsification of activations and gradients significantly reduces the communication overhead. 
Empirical study have suggested that the proposed framework guarantees a low accuracy loss, while only requiring $0.2\%$ of the synchronization traffic in vanilla federated learning.   

\end{abstract}

\section{Introduction}
Machine learning is bringing a paradigm shift to healthcare, powered by increasing availability of healthcare data and rapid progress of analytic techniques \cite{jiang2017artificial}. 
Among the large volume of healthcare data, the internet of things (IoT) have become one of the most important data sources. 
In fact, IoT has shown great potential in many medical applications such as remote health monitoring, fitness programs, chronic diseases, and elderly care \cite{islam2015internet}. 
One advantage of IoT technologies is to support systems capable of continuously clinical-level monitoring of subjects' conditions and acquiring a variety of bio-signals. 
As a result, an ample range of applications for individualized eHealth technologies have emerged.

Traditionally, data collected by IoT devices are uploaded to a data center and further leveraged to train machine learning models. However, data owners are increasingly privacy sensitive \cite{lim2020federated}, especially in the healthcare domain, to meet the privacy requirement regarding individually identifiable health information (eg., HIPAA \cite{health2018nccl}). Federated learning, first proposed by Google \cite{konevcny2016federated, mcmahan2017communication}, is a new attempt to resolve this data dilemma, which is defined as a problem of training a high-quality shared global model with a central server from decentralized data scattered among a large number of clients \cite{xu2019federated}.

Unfortunately, vanilla federated learning is not sufficient to meet the requirement of healthcare IoT devices which typically have limited energy storage, low computational capacity, and restricted network bandwidth compared with standard mobile devices \cite{firouzi2018internet}. Inspired by SplitNN \cite{vepakomma2018split, singh2019detailed}, which is initially designed to avoid sharing raw data or model details with collaborating medical institutions, we decompose a deep neural network into two parts: i) a very shallow sub-network in the IoT device that processes client's data; ii) a deep sub-network in the cloud that contains most of the computation load of the original model. Then, we further reduce the communication overhead during training by sparsifying the tensor when transferring activations and gradients for the forward and backward propagation.  
Empirically, we show that the proposed framework can preserve a very small accuracy loss and significantly reduce the communication overhead for a state of the art architecture to detect arrhythmia detection by monitoring single lead electrocardiograms \cite{hannun2019cardiologist}.

\section{Decomposed Federated Learning}

We first review important relevant concepts, and then formally define how to decompose the neural network for computation and sparsify activations and gradients for communication. 

\textbf{Multiple Layer Neural Network.} A deep neural network is designed to approximate a target function ${\bf y}=f^*\left({\bf x}\right)$, which maps an input feature $\bf x$ to output prediction $\bf y$. Formally, the function $f^*$ is composed by a chain of $N$ different functions as: $f^*\left({\bf x}\right)=f^{N}\left(f^{N-1}...\left(f^{2}\left(f^{1}\left({\bf x}\right)\right)\right)\right)$.

\textbf{Federated Learning.}
A classic federated learning systems includes $M$ data owners who need to train models $\left\{f_1,f_2,...,f_M\right\}$ on their respective datasets $\left\{D_1,D_2,...,D_M\right\}$. According to the Categorization in \cite{yang2019federated}, the IoT healthcare scenario falls under the category of horizontal federated learning, where datasets share the same feature space but differ in samples. Formally, the aim is to minimize $f\left(x\right)$ w.r.t., parameter $w$: 
\begin{equation*}
\underset{w}{\min} \space f\left(x\right) = \sum_{j=1}^{M}f_j\left(w\mid D_j\right)
\end{equation*}
\textbf{Decompose the Neural Network.} Inspired by model parallelism \cite{chilimbi2014project}, eg., splitNN \cite{vepakomma2018split}, we decompose the approximated function $f^*$ so that each IoT device (indexed by $j$) will include a local version of the first shallow component ${\bf a}^1=f^1_j\left({\bf x}\right)$, while the rest part ${\bf y}=f^{N}\left(...\left(f^{2}\left({{\bf a}^1}\right)\right)\right)$ is allocated on the centralized server. We can formalize the new aim as:
\begin{equation*}
\underset{w}{\min} \space f\left(x\right) = \sum_{j=1}^{M}f^{N}\left(...\left(f^{2}\left(w^{2,...,N}\mid f^1_j\left(w^1_j\mid D_j\right)\right)\right)\right)
\end{equation*}
Note that such partition will dramatically decrease the computation load on each edge device while preserving the data privacy constraint.

\textbf{Sparsify Activations and Gradients.} Different from model synchronization in vanilla federated learning, the proposed framework needs to communication the activations ${\bf a}^1$ in the forward propagation and the corresponding gradients $d{\bf a}^1$ in the backward propagation. 
To further reduce the network traffic, we extend the idea of sparsification of gradients \cite{stich2018sparsified, alistarh2018convergence}. 
To be specific, we sparsify ${\bf a}^1$, $d{\bf a}^1$ by only communicating the top K$(\leq10\%)$  elements at each iteration. 

\section{Empirical Study}
We consider using this proposed framework to detect arrhythmia by monitoring single lead electrocardiograms.  We apply the state of the art architecture \cite{hannun2019cardiologist} on the PhysioNet 2017 dataset \cite{physionet2017}.

\textbf{Experiment Setup.} We include $74275$ segments of ECG signal in the training set and $13107$ segments in the test set. We only include the signals labeled of arrhythmia and normal, and split the original signal into segments of length $256$. In this preliminary experiment, we focus on two aspects: i) how the proposed framework converges comparing to vanilla SGD in a centralized server; ii) the network traffic reduction of the proposed framework comparing with other federated learning algorithms.

\begin{figure}[t]
	\centering
	\includegraphics[width=0.42\textwidth]{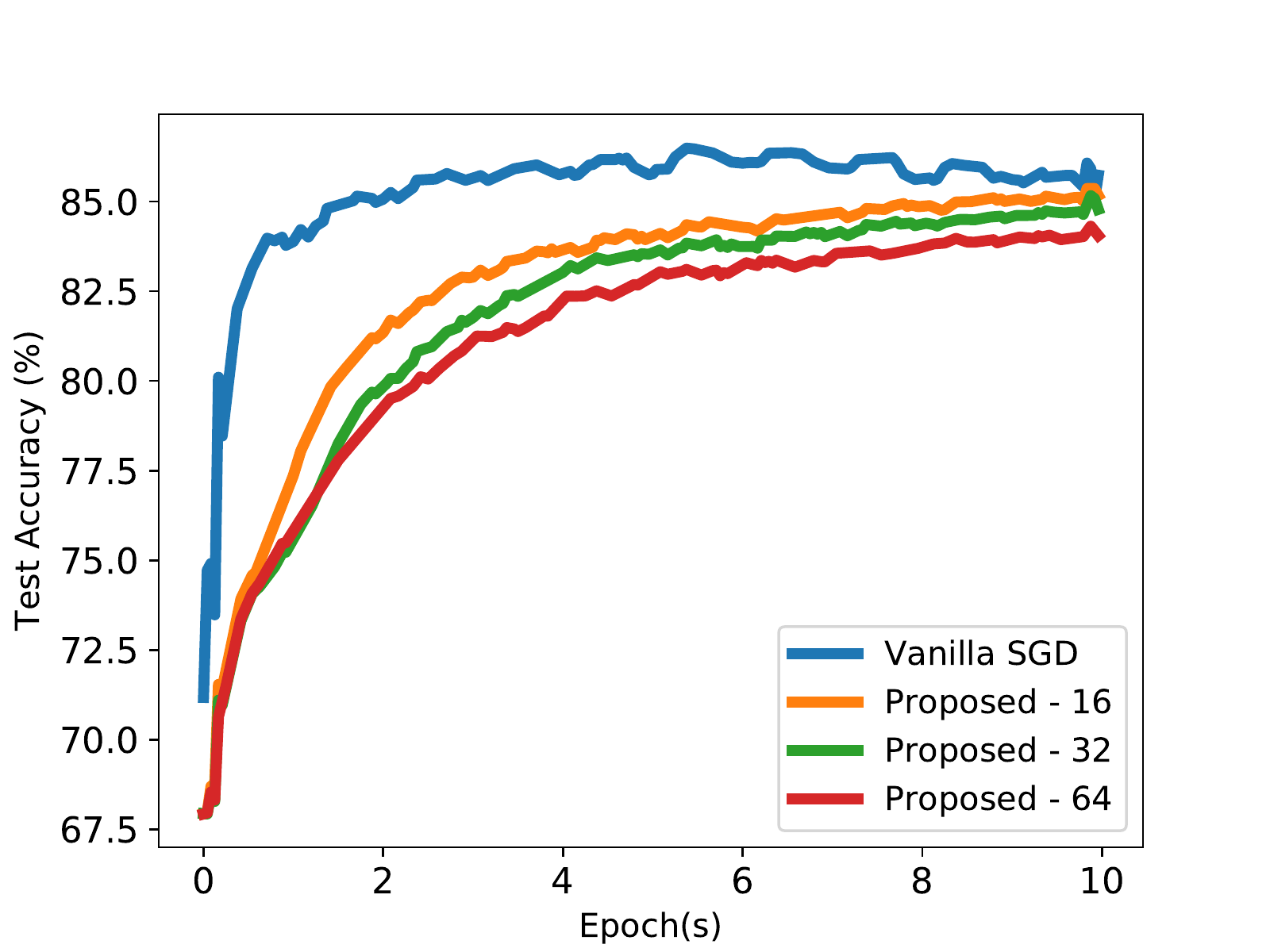}
	\caption{Convergence of ResNet in PhysioNet 2017.}
	\label{figure_plot}
\end{figure}

\textbf{Convergence and Accuracy Loss.} Figure \ref{figure_plot} shows the convergence comparison between vanilla SGD and the proposed algorithm over $16$, $32$, and $64$ IoT devices. We observe that there is only an insignificant delay of the convergence, while the final accuracy loss is less than $2\%$. 

\textbf{Communication Overhead Reduction.} Table \ref{table_scalability} illustrates the network traffic required for a single iteration (with a batch size of $32$ on each device) over FedAvg \cite{mcmahan2017communication}, SplitNN \cite{vepakomma2018split}, and the proposed framework. Note that we assign the first convolution layer in the edge device for both SplitNN and the proposed framework. As it reveals in Table \ref{table_scalability}, the proposed algorithm reduces $99.8\%$, $90\%$ network traffic comparing to FedAvg and SplitNN respectively.  
\begin{table}[]
\centering
\begin{tabular}{@{}c|c|c|c@{}} \hline
           & FedAvg & SplitNN & Proposed \\ \hline
16 devices & 1.36GB & 32MB    & 3.2MB    \\ 
32 devices & 2.72GB & 64MB    & 6.4MB    \\ 
64 devices & 5.45GB & 128MB   & 12.8MB   \\ \hline
\end{tabular}
\caption{Network traffic comparison for each iteration.}
	\label{table_scalability}
\end{table}

\section{Conclusion and Future Work}{\label{sec:conclusion}}
We propose a novel federated learning framework for healthcare IoT devices, where we significantly reduce the computation load on the IoT devices, and communication overhead between the IoT devices and the centralized server. Additionally, we observe very small accuracy loss in a real world  arrhythmia detection task. 

On the other hand, this proposal also leaves a few open questions: i) how to provide theoretical analysis of accuracy loss upper-bound and guarantee of convergence w.r.t., activation and gradient sparasification; ii) how to design a more comprehensive system that can simultaneously manage multiple learning tasks on multi-sensor healthcare IoT devices. We hope this proposal will shed light on these interesting directions in privacy sensitive edge computing for healthcare machine learning applications. 


\bibliography{main}
\bibliographystyle{icml2020}

\end{document}